\documentclass{article}

\usepackage{PRIMEarxiv}

\usepackage[utf8]{inputenc} 
\usepackage[T1]{fontenc}    
\usepackage{hyperref}       
\usepackage{url}            
\usepackage{booktabs}       
\usepackage{amsfonts}       
\usepackage{nicefrac}       
\usepackage{microtype}      
\usepackage{lipsum}
\usepackage{fancyhdr}       
\usepackage{graphicx}       
\graphicspath{{media/}}     

\usepackage{url,hyperref,lineno,microtype,subcaption}
\usepackage{multirow}
\usepackage{amsmath}

\title{An Integrated Framework for Multi-Granular Explanation of Video Summarization
\thanks{\textit{\underline{Citation}}: 
\textbf{K. Tsigos, E. Apostolidis, V. Mezaris. An Integrated Framework for Multi-Granular Explanation of Video Summarization. Under Review.}} 
}

\author{
  Konstantinos Tsigos, Evlampios Apostolidis, Vasileios Mezaris \\
  Information Technologies Institute (ITI), Centre for Research and Technology, Hellas (CERTH) \\
  Thessaloniki, Greece\\
  \texttt{\{ktsigos, apostolid, bmezaris\}@iti.gr} \\
}

\begin{document}
\maketitle

\begin{abstract}
In this paper, we propose an integrated framework for multi-granular explanation of video summarization. This framework integrates methods for producing explanations both at the fragment level (indicating which video fragments influenced the most the decisions of the summarizer) and the more fine-grained visual object level (highlighting which visual objects were the most influential for the summarizer). To build this framework, we extend our previous work on this field, by investigating the use of a model-agnostic, perturbation-based approach for fragment-level explanation of the video summarization results, and introducing a new method that combines the results of video panoptic segmentation with an adaptation of a perturbation-based explanation approach to produce object-level explanations. The performance of the developed framework is evaluated using a state-of-the-art summarization method and two datasets for benchmarking video summarization. The findings of the conducted quantitative and qualitative evaluations demonstrate the ability of our framework to spot the most and least influential fragments and visual objects of the video for the summarizer, and to provide a comprehensive set of visual-based explanations about the output of the summarization process.
\end{abstract}


\section{Introduction}
The current practice in the Media industry for producing a video summary requires a professional video editor to watch the entire content and decide about the parts of it that should be included in the summary. This is a laborious task and can be very time-consuming in the case of long videos or when different summaries of the same video should be prepared for distribution via multiple video sharing platforms (e.g., YouTube, Vimeo, TikTok) and social networks (e.g., Facebook, Twitter, Instagram) with different specifications about the optimal or maximum video duration \cite{10.1007/978-3-031-53302-0_21}. Video summarization technologies aim to generate a short summary by selecting the most informative and important frames (key-frames) or fragments (key-fragments) of the full-length video, and presenting them in temporally-ordered fashion. The use of such technologies by Media organizations can drastically reduce the needed resources for video summarization in terms of both time and human effort, and facilitate indexing, browsing, retrieval and promotion of their media assets \cite{apostolidis_chapter}. Despite the recent advances in the field of video summarization, which are tightly associated with the emergence of modern deep learning network architectures \cite{9594911}, the outcome of a video summarization method still needs to be curated by a video editor, to make sure that all the necessary parts of the video were included in the summary. This content production step could be further facilitated if the video editor is provided with explanations about the suggestions made by the used video summarization technology. The provision of such explanations would allow a level of understanding about the functionality of this technology, thus increasing the editor's trust in it and facilitating content curation. 

Over the last years there is an increasing interest in explaining the outcomes of deep networks processing video data. Nevertheless, most works are related with network architectures for video classification \cite{Bargal_2018_CVPR, 10.1007/978-3-030-69541-5_25, Li2021TowardsVE}, action classification and reasoning \cite{8803153, 10.1145/3343031.3351040, HAN2022212}, activity recognition \cite{Aakur2018AnIE, Roy2019ExplainableAR} and anomaly detection \cite{278c9656de614a479c93c6dead189ff4, 10205367, 9468958, 9706981, 8237653}. With respect to explainable video summarization, a first attempt to formulate the task and evaluate various attention-based explanation signals was initially reported in \cite{10019643} and extended in \cite{10.1145/3607540.3617138}. Another approach that relies on the use of causality graphs between input data, output scores, summarization criteria and data perturbations, was presented in \cite{10208771}. However, the produced graphs require interpretation by a human expert, while the performance of these explanations was not evaluated through quantitative or qualitative analysis.

In this paper, we build on our previous efforts on explainable video summarization \cite{10019643, 10.1145/3607540.3617138} and extend them, in order to: i) investigate the use of a model-agnostic approach (adaptation of the LIME method \cite{ribeiro2016should}) for fragment-level explanation of the video summarization results, ii) develop a new method for producing more fine-grained explanations at the visual object level that provide more insights about the focus of the summarizer, and iii) build an integrated framework for multi-granular (and thus more informative) explanation of the video summarization results. Our contributions are the following:
\begin{itemize}
    \item We adapt the model-agnostic LIME explanation method \cite{ribeiro2016should} to operate on sequences of video frames (rather than on a single frame/image, which is the typical approach) and produce a fragment-level explanation of the video summarization results, which indicates the temporal fragments of the video that influenced the most the decisions of the summarizer.
    \item We combine the state-of-the-art Video K-Net method for video panoptic segmentation \cite{li2022videoknet} with another adaptation of the LIME method \cite{ribeiro2016should} that also operates on frame sequences, to build a method that performs object-oriented perturbations over a sequence of frames and produces explanations at the level of visual objects.
    \item We integrate the methods for fragment- and object-level explanation into a framework for multi-granular explanation of video summarization, and assess their performance based on quantitative and qualitative evaluations using a state-of-the-art method (CA-SUM \cite{10.1145/3512527.3531404}) and two datasets for video summarization (SumMe \cite{10.1007/978-3-319-10584-0_33} and TVSum \cite{7299154}).
\end{itemize}

\section{Related Work}

\begin{table}[t]
\centering
\caption{Surveyed works on explainable video analysis. For each work we outline the adopted explanation approach and the targeted task.}
\label{tab:soa}
\begin{tabular}{|l|l|l|}
\hline
\textbf{Work}      & \textbf{Explanation approach}                                                                                                                               & \textbf{Video analysis task}                                                                        \\ \hline
Bargal et al. (2018) \cite{Bargal_2018_CVPR}      & \begin{tabular}[c]{@{}l@{}}Use of spatio-temporal cues to spot fragments\\ linked to specific action/phrase from caption\end{tabular}           & \begin{tabular}[c]{@{}l@{}}Classification\\ \& captioning\end{tabular}         \\ \hline
M{\"a}ntt{\"a}ri et al. (2020) \cite{10.1007/978-3-030-69541-5_25}    & \begin{tabular}[c]{@{}l@{}}Perturbation-based detection of the most\\ influential video fragment\end{tabular}                                       & Classification                                                                 \\ \hline
Li et al. (2021) \cite{Li2021TowardsVE}          & \begin{tabular}[c]{@{}l@{}}Perturbation-based method for spatio-temporally\\ smooth explanation\end{tabular}                                    & Classification                                                                 \\ \hline
Stergiou et al. (2019) \cite{8803153}    & Use of heatmaps visualizing the focus of attention                                                                                                  & \begin{tabular}[c]{@{}l@{}}Action classification\\ \& recognition\end{tabular} \\ \hline
Zhuo et al. (2019) \cite{10.1145/3343031.3351040}       & \begin{tabular}[c]{@{}l@{}}Use of spatio-temporal graph of semantic-level \\ video states\end{tabular}                                          & Action reasoning                                                               \\ \hline
Han et al. (2022) \cite{HAN2022212}         & \begin{tabular}[c]{@{}l@{}}Target-aware tracking strategy to estimate objects'\\ temporal relevance \& form a scene graph\end{tabular}          & Action reasoning                                                               \\ \hline
Aakur et al. (2018) \cite{Aakur2018AnIE}       & \begin{tabular}[c]{@{}l@{}}Use of connected structures of the detected visual\\ concepts to form explainable representations\end{tabular}       & Activity recognition                                                           \\ \hline
Roy et al. (2019) \cite{Roy2019ExplainableAR}         & \begin{tabular}[c]{@{}l@{}}Use of a tractable interpretable probabilistic\\ graphical model\end{tabular}                                        & Activity recognition                                                           \\ \hline
Wu et al. (2022) \cite{278c9656de614a479c93c6dead189ff4}          & \begin{tabular}[c]{@{}l@{}}Extract high-level concept \& context features\\ to train a denoising autoencoder\end{tabular}                       & Anomaly detection                                                              \\ \hline
Guo et al. (2022) \cite{9468958}         & \begin{tabular}[c]{@{}l@{}}Visualization tool for comparing normal and\\ abnormal sequences in a latent space\end{tabular}                      & Anomaly detection                                                              \\ \hline
Szymanowicz et al. (2022) \cite{9706981} & \begin{tabular}[c]{@{}l@{}}Use of saliency maps to provide spatial location\\ and representation of the anomalous event\end{tabular}            & Anomaly detection                                                              \\ \hline
Hinami et al. (2017) \cite{8237653}      & \begin{tabular}[c]{@{}l@{}}Compute semantic anomaly scores using a\\ context-sensitive anomaly detector\end{tabular}                            & Anomaly detection                                                              \\ \hline
Singh et al. (2023) \cite{10205367}       & \begin{tabular}[c]{@{}l@{}}Use of learned representations of the depicted\\ objects and their motions\end{tabular}                              & Anomaly localization                                                           \\ \hline
Papoutsakis et al. (2019) \cite{Papoutsakis2019UnsupervisedAE} & \begin{tabular}[c]{@{}l@{}}Use action graphs representing objects and\\ behaviors\end{tabular}                                                  & Similarity evaluation                                                          \\ \hline
Gkalelis et al. (2022) \cite{9915576}    & \begin{tabular}[c]{@{}l@{}}Use of weighted in-degrees of graph attention\\ networks’ adjacency matrices\end{tabular}                            & Event recognition                                                              \\ \hline
Yu et al. (2021) \cite{YU2021107791}          & \begin{tabular}[c]{@{}l@{}}Trainable framework combining spatial \& motion\\ information with appearance-geometry descriptor\end{tabular}       & Text detection                                                                 \\ \hline
Apostolidis et al. (2022) \cite{10019643} & \begin{tabular}[c]{@{}l@{}}Use of attention weights to form video-fragment-\\ level explanations\end{tabular}                                   & Summarization                                                                  \\ \hline
Huang et al. (2023) \cite{10208771}       & \begin{tabular}[c]{@{}l@{}}Causality graphs of input data, output scores,\\ summarization criteria and data perturbations.\end{tabular} & Summarization                                                                  \\ \hline
\end{tabular}
\end{table}

Over the last years there is a rapidly growing interest of researchers on building methods that provide explanations about the working mechanism or the decisions/predictions of neural networks. Nevertheless, in contrast to the notable progress in the fields of pattern recognition \cite{BAI2021108102}, image classification \cite{10.1007/978-3-031-25085-9_23, ntrougkas2024ttame}, and NLP \cite{10.1145/3529755}, currently there are only a few works on producing explanations for networks that process video data (listed in Table \ref{tab:soa}). Working with network architectures for video classification, Bargal et al. (2018) \cite{Bargal_2018_CVPR} visualized the spatio-temporal cues contributing to the network’s classification/captioning output using internal representations and employed these cues to localize video fragments corresponding to a specific action or phrase from the caption. M{\"a}ntt{\"a}ri et al. (2020) \cite{10.1007/978-3-030-69541-5_25} utilized the concept of meaningful perturbation to spot the video fragment with the greatest impact on the video classification results. Li et al. (2021) \cite{Li2021TowardsVE} extended a generic perturbation-based explanation method for video classification networks by introducing a loss function that constraints the smoothness of explanations in both spatial and temporal dimensions. Focusing on methods for action classification and reasoning, Stergiou et al. (2019) \cite{8803153} proposed the use of cylindrical heat-maps to visualize the focus of attention at a frame basis and form explanations of deep networks for action classification and recognition. Zhuo et al. (2019) \cite{10.1145/3343031.3351040} defined a spatio-temporal graph of semantic-level video states (representing associated objects, attributes and relationships) and applied state transition analysis for video action reasoning. Han et al. (2022) \cite{HAN2022212} presented a one-shot target-aware tracking strategy to estimate the relevance between objects across the temporal dimension and form a scene graph for each frame, and used the generated video graph (after applying a smoothing mechanism) for explainable action reasoning. Dealing with networks for video activity recognition, Aakur et al. (2018) \cite{Aakur2018AnIE} formulated connected structures of the detected visual concepts in the video (e.g., objects and actions) and utilized these structures to produce semantically coherent and explainable representations for video activity interpretation, while Roy et al. (2019) \cite{Roy2019ExplainableAR} fed the output of a model for activity recognition to a tractable interpretable probabilistic graphical model and performed joint learning over the two. In the field of video anomaly detection, Wu et al. (2022) \cite{278c9656de614a479c93c6dead189ff4} extracted high-level concept and context features for training a denoising autoencoder that was used for explaining the output of anomaly detection in surveillance videos. Guo et al. (2022) \cite{9468958} constructed a sequence-to-sequence model (based on a variational autoencoder) to detect anomalies in videos and combined it with a visualization tool that facilitates comparisons between normal and abnormal sequences in a latent space. Szymanowicz et al. (2022) \cite{9706981} designed an encoder-decoder architecture to detect anomalies, that is based on U-Net \cite{10.1007/978-3-319-24574-4_28}, thereby generating saliency maps by computing per-pixel differences between actual and predicted frames; based on the per-pixel squared errors in the saliency maps, they introduced an explanation module that can provide spatial location and human-understandable representation of the identified anomalous event. Hinami et al. (2017) \cite{8237653} employed a Fast R-CNN-based model to learn multiple concepts in videos and extract semantic features, and applied a context-sensitive anomaly detector to obtain semantic anomaly scores which can be seen as explanations for anomalies. Singh et al. (2023) \cite{10205367} developed an explainable method for single-scene video anomaly localization, which uses learned representations of the depicted objects and their motions to provide justifications on why a part of the video was classified as normal or anomalous. Working with network architectures that tackle other video analysis and understanding tasks, Papoutsakis et al. (2019) \cite{Papoutsakis2019UnsupervisedAE} presented an unsupervised method that evaluates the similarity of two videos based on action graphs representing the detected objects and their behavior, and provides explanations about the outcome of this evaluation. Gkalelis et al. (2022) \cite{9915576} used the weighted in-degrees of graph attention networks' adjacency matrices to provide explanations of video event recognition, in terms of salient objects and frames. Yu et al. (2021) \cite{YU2021107791} built an end-to-end trainable and interpretable framework for video text detection with online tracking that captures spatial and motion information and uses an appearance-geometry descriptor to generate robust representations of text instances. Finally, a few attempts were made towards explaining video summarization. Apostolidis et al. (2022, 2023) \cite{10019643,10.1145/3607540.3617138} formulated the task as the production of an explanation mask indicating the parts of the video that influenced the most the estimates of a video summarization network about the frames' importance. Then, they utilized a state-of-the-art network architecture (CA-SUM \cite{10.1145/3512527.3531404}) and two datasets for video summarization (SumMe \cite{10.1007/978-3-319-10584-0_33} and TVSum \cite{7299154}), and evaluated the performance of various attention-based explanation signals by investigating the network's input-output relationship (according to different input replacement functions), and using a set of tailored evaluation measures. Following a different approach, Huang et al. (2023) \cite{10208771} described a method for explainable video summarization that leverages ideas from Bayesian probability and causation modeling. A form of explanation about the outputs of this method is provided through causality graphs that show relations between input data, output importance scores, summarization criteria (e.g., representativeness, interestingness) and applied perturbations.

Differently to most of the above discussed works that deal with the explanation of network architectures trained for various video analysis tasks (e.g., classification, action and activity recognition, anomaly detection), in this work, we focus on networks for video summarization. Contrary to the work of \cite{10208771}, our framework produces visual-based explanations indicating the parts of the video (temporal video fragments and visual objects) that influenced the most the decisions of the summarizer, rather than providing causality graphs that need interpretation by a human expert. Moreover, the performance of our framework is assessed through a set of quantitative and qualitative evaluations. As stated in the introduction, our work builds on our previous efforts for explaining video summarization \cite{10019643, 10.1145/3607540.3617138} and extends them by: i) examining the use of a model-agnostic approach for producing fragment-level explanations (rather than requiring access to the internal layers and weights of the summarization network), ii) proposing a novel methodology for producing object-level explanations (thus providing more clues about the content of the video that is more important for the summarizer), and iii) combining the different explanation approaches under an integrated framework that offers a multi-granular, and thus more comprehensive explanation for the output of the video summarization process.

\begin{figure}[t]
\centering
\includegraphics[width=\textwidth]{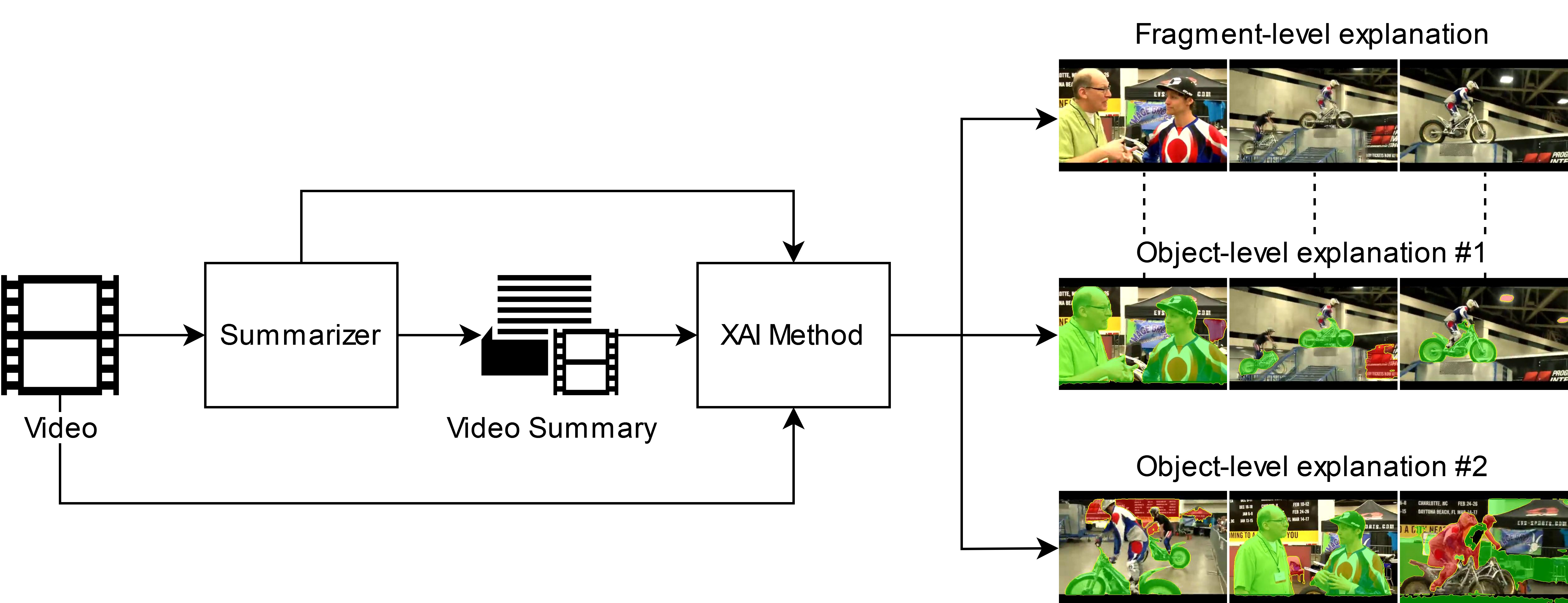}
\caption{High-level overview of our framework for explaining video summarization. This framework produces: i) a fragment-level explanation indicating the most influential video fragments; ii) object-level explanation (\#1) highlighting the most influential objects within the most influential fragments; and iii) object-level explanation (\#2) highlighting the visual objects within the fragments that have been selected for inclusion in the summary, that influenced the most this selection.}
\label{fig:high-level}
\end{figure}

\section{Proposed Approach}

A high-level overview of the developed framework for multi-granular explainable video summarization is given in Fig. \ref{fig:high-level}. Given an input video, a summarizer and the produced video summary (formed by the three top-scoring video fragments by the summarizer), our framework produces three different types of explanations: i) a fragment-level explanation that indicates the temporal video fragments that influenced the most the decisions of the summarizer, ii) an object-level explanation that highlights the most influential visual objects within the aforementioned fragments, and iii) another object-level explanation that points out the visual objects within the fragments that have been selected for inclusion in the summary, that influenced the most this selection. In the core of this framework there is an XAI method that is responsible for producing the explanation. More details about the processing steps and the employed XAI method for producing each type of explanation, are provided in the following sections. 

\begin{figure}[t]
\centering
\includegraphics[width=0.88\textwidth]{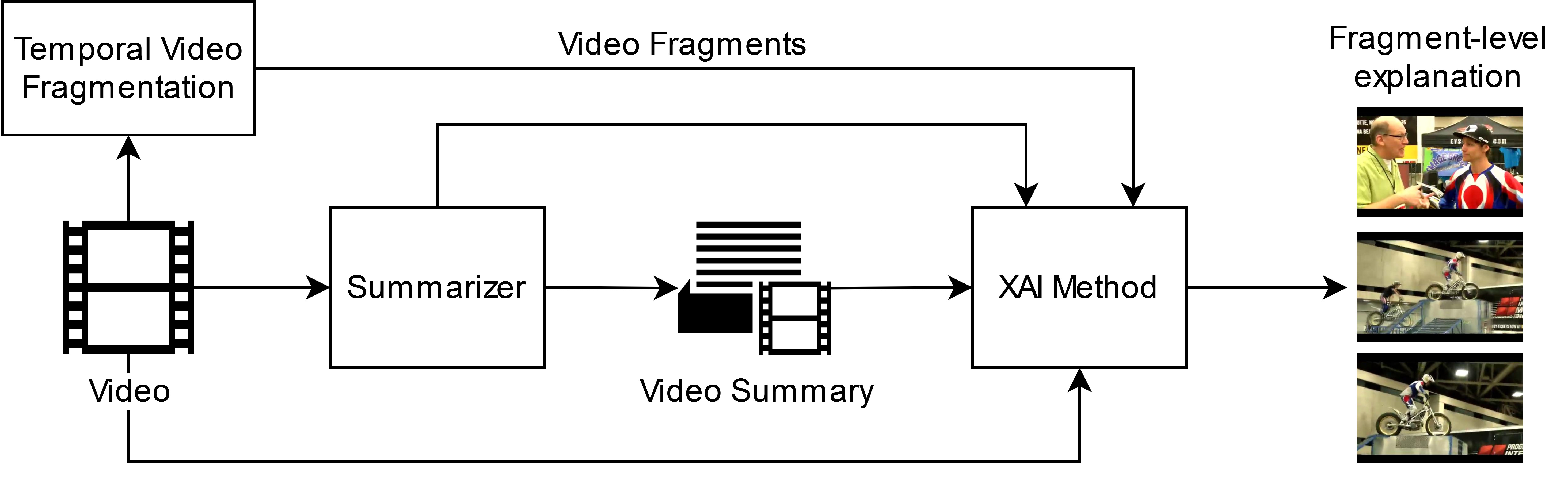}
\caption{The proposed processing pipeline for producing fragment-level explanations.}
\label{fig:fragment-level}
\end{figure}

\subsection{Fragment-level Explanation}

The processing pipeline for producing fragment-level explanations is depicted in Fig. \ref{fig:fragment-level}. As can be seen, the input video needs to be temporally fragmented into consecutive and non-overlapping fragments. To perform this process, we employ a pre-trained model of the TransNetV2 method for shot segmentation from \cite{soucek2020transnetv2}. This method relies on a 3D-CNN network architecture with two prediction heads; one predicting the middle frame of a shot transition and another one predicting all transition frames and used during training to improve the network's understanding of what constitutes a shot transition and how long it is. The used model has been trained using synthetically-created data from the TRECVID IACC.3 dataset \cite{DBLP:conf/trecvid/AwadBFJDMSGJKQE17} and the ground-truth data of the ClipShots dataset \cite{10.1007/978-3-030-20887-5_36}. If the number of video fragments is equal to one (thus, the input video is a single-shot user-generated video) or less than ten (thus, the selection of three fragments for building the summary would not lead to a significantly condensed synopsis of the video), we further fragment the input video using the method for sub-shot segmentation from \cite{apostolidis2018motion}. This method segments a video into visually coherent parts that correspond to individual video capturing activities (e.g., camera pan and tilt, change in focal length and camera displacement) by extracting and evaluating the region-level spatio-temporal distribution of the optical flow over sequences of neighbouring video frames.

The defined video fragments along with the input video, the summarizer and the produced video summary, are then given as input to the XAI method. This method can be either model-agnostic (i.e., it does not require any knowledge about the summarization model) or model-specific (i.e., it utilizes information from the internal layers of the model). In this work, we considered the LIME explanation method from \cite{ribeiro2016should} and the best-performing configuration of the attention-based explanation method from \cite{10019643}, respectively. LIME \cite{ribeiro2016should} is a perturbation-based method that approximates the behavior of a model locally by generating a simpler, interpretable model. This method was designed for producing image-level explanations by masking out regions of the image; thus, we had to adapt it to operate over sequences of frames and produce fragment-level explanations. In particular, instead of masking out regions of a video frame during a perturbation, we mask out entire video fragments by replacing their frames with black frames. The perturbed version of the input video is fed to the summarizer, which then produces a new output (i.e., a new sequence of frame-level importance scores). This process is repeated $M$ times and the binary masks of each perturbation (indicating the fragments of the video that were masked out) are fitted to the corresponding importance scores (computed as the average of the sequence of frame-level importance scores) using a linear regressor. Finally, the fragment-level explanation is produced by focusing on the top-3 scoring fragments (indicated by the assigned weights to the indices of the binary masks) by this simpler model. The attention-based method of \cite{10019643} can be applied on network architectures for video summarization that estimate the frames' importance with the help of an attention mechanism, such as the ones from \cite{10.1145/3512527.3531404,10.1007/978-3-030-21074-8_4,LI2021107677}. This method uses the computed attention weights in the main diagonal of the attention matrix for a given input video, and forms an explanation signal by averaging them at the fragment level. The values of this explanation signal indicate the influence of the video's fragments in the output of the summarizer, and the fragments related to the top-3 scoring ones are selected to create the fragment-level explanation.

\begin{figure}[t]
\centering
\includegraphics[width=0.73\textwidth]{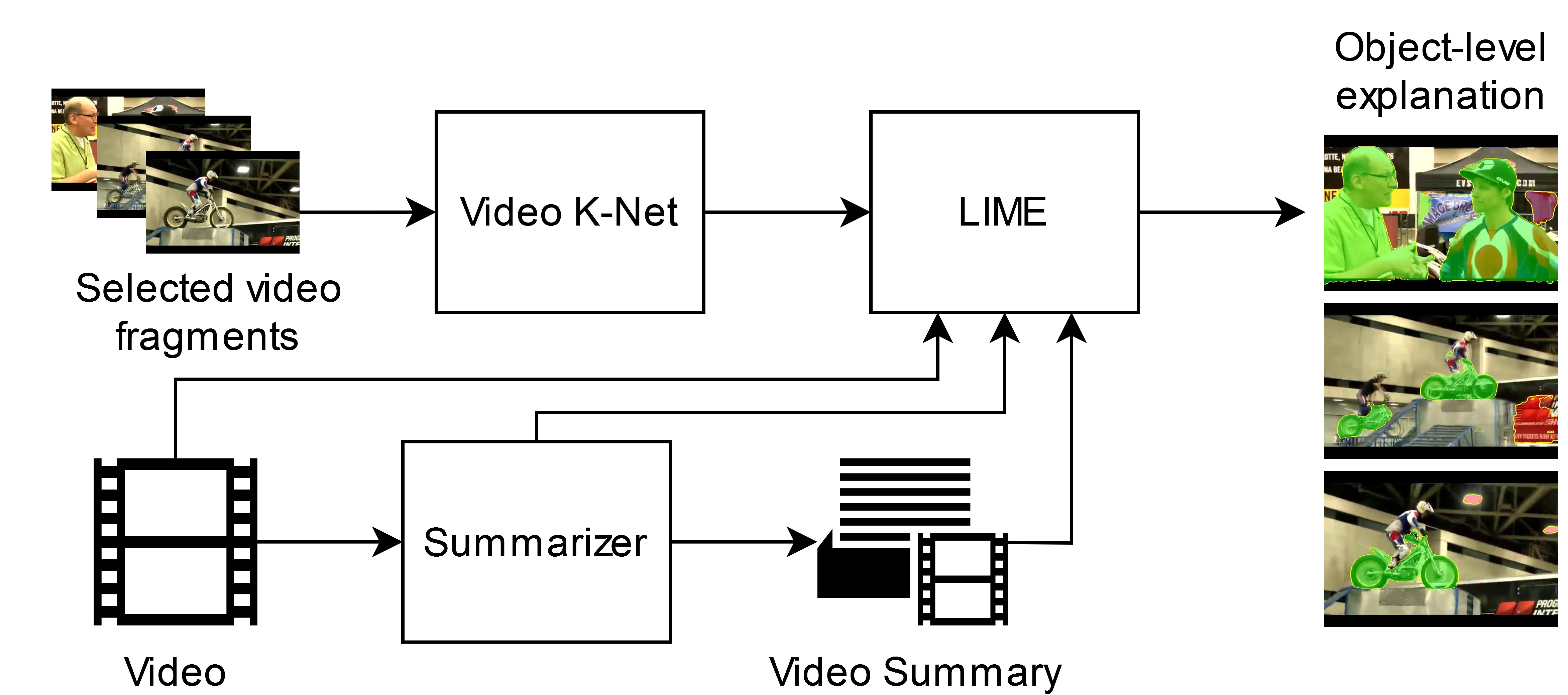}
\caption{Processing pipeline for producing object-level explanations. The selected video fragments are the most influential according to the fragment-level explanation, or the top-scoring by the summarizer.}
\label{fig:vis-level}
\end{figure}

\subsection{Object-level Explanation}

The processing pipeline for creating object-level explanations is shown in Fig. \ref{fig:vis-level}. The selected video fragments for creating such explanations can be either the most influential ones according to the fragment-level explanation, or the top-scoring ones by the summarizer, that were selected for inclusion in the video summary. The utilized XAI method in this case is LIME \cite{ribeiro2016should}, and the goal is to apply perturbations at the visual object level in order to identify the objects within the selected fragments, that influence the most the output of the summarizer. Once again, we use an adaptation of the LIME method, that takes into account the applied spatial perturbations in the visual content of a sequence of video frames (and not on a single frame). To make sure that a perturbation is applied on the same visual object(s) across the frames of a video fragment, we spatially segment these frames using a model of the Video K-Net method for video panoptic segmentation \cite{li2022videoknet}, trained on the VIP-Seg dataset \cite{miao2022large}. This method builds about the foundation of K-Net \cite{zhang2021k}, which unifies image segmentation through a collection of adaptable kernels. Video K-Net capitalizes on the kernels' ability to encode object appearances and contextual information, combining segmentation and tracking of both semantically meaningful categories and to individual instances of countable objects across a sequence of video frames. The top-scoring frame (by the summarizer) within a selected video fragment (by the fragment-level explanation or the summarizer) is picked as the keyframe. Once all the frames of this fragment have been spatially segmented by Video K-Net, the appearing visual objects in the selected keyframe are masked out across the entire video fragment through a series of perturbations that replace the associated pixels of the video frames (specified by the assigned object IDs from the Video K-Net method) with black pixels. The perturbed version of the input video after masking out a visual object in one of the selected video fragments is forwarded to the summarizer, which outputs a new sequence of frame-level importance scores. This process is repeated $N$ times for a given video fragment and the binary masks of each perturbation are fitted to the corresponding importance scores (computed as the average of the importance scores of the frames within the selected fragment) using a linear regressor. Finally, the object-level explanation is formed by taking the top- and bottom-scoring visual objects (indicated by the assigned weights to the indices of the binary masks) by this simpler model, and highlighting the corresponding visual objects (using green and red coloured overlaying masks, respectively) in the keyframes of the selected video fragments.

\section{Experiments}

In this section we describe the utilized datasets and evaluation protocol for assessing the performance of the produced explanations. Following, we provide some implementation details and report the findings of the conducted quantitative and qualitative evaluations.

\subsection{Datasets and Evaluation Protocol}
\label{subsec:data_eval}

In our experiments we employ the SumMe \cite{10.1007/978-3-319-10584-0_33} and TVSum \cite{7299154} datasets, which are the most widely used ones in the literature for video summarization \cite{9594911}. SumMe is composed of $25$ videos with diverse video contents (e.g., covering holidays, events and sports), captured from both first-person and third-person view. TVSum contains $50$ videos from $10$ categories of the TRECVid MED task. For evaluation we utilize the Discoverability+/- and Sanity Violation measures from \cite{10019643}. The Rank Correlation measure was not taken into account, as we are interested in the capacity of explanations to spot the most and least influential fragments, rather than ranking the entire set of video fragments based on their influence to the summarization method. For completeness, in the following we describe each measure and the way it was computed in our evaluations.

To measure the influence of a selected video fragment or visual object by an explanation method, we mask it out (using black frames or pixels, respectively) and compute the difference in the summarization model's output, as $\Delta E(\boldsymbol{X},\boldsymbol{\hat{X}^{k}}) = \tau(\boldsymbol{y}, \boldsymbol{y^{k}})$. In this formula, $\boldsymbol{X}$ is the set of original frame representations, $\boldsymbol{\hat{X}^{k}}$ is the set of updated features of the frames belonging to the selected $k^{th}$ video fragment (after the applied mask out process), $\boldsymbol{y}$ and $\boldsymbol{y^{k}}$ are the outputs of the summarization model for $\boldsymbol{X}$ and $\boldsymbol{\hat{X}^{k}}$, respectively, and $\tau$ is the Kendall's $\tau$ correlation coefficient \cite{kendall1945treatment}. Based on $\Delta E$, we assess the performance of each explanation using the following evaluation measures:
\begin{itemize}
    \item \textbf{Discoverability+ (Disc+)} evaluates if the top-3 scoring fragments/objects by an explanation method have a significant influence to the model's output. For a given video, it is calculated by computing $\Delta E$ after perturbing (masking out) the top-1, top-2 and top-3 scoring fragments/objects in a one-by-one and sequential (batch) manner. The higher this measure is, the greater the ability of the explanation to spot the video fragments or visual objects with the highest influence to the summarization model.
    \item \textbf{Discoverability- (Disc-)} evaluates if the bottom-3 scoring fragments/objects by an explanation method have small influence to the model's output. For a given video, it is calculated by computing $\Delta E$ after perturbing (masking out) the bottom-1, bottom-2 and bottom-3 scoring fragments/objects in a one-by-one and sequential (batch) manner. The lower this measure is, the greater the effectiveness of the explanation to spot the video fragments or visual objects with the lowest influence to the summarization model.
    \item \textbf{Sanity Violation (SV)} quantifies the ability of explanations to correctly discriminate the most from the least influential video fragments or visual objects. It is calculated by counting the number of cases where the condition (Disc+ $>$ Disc-) is violated, after perturbing (masking out) parts of the input corresponding to fragments/objects with the three highest and lowest explanation scores in a one-by-one and sequential (batch) manner, and then expressing the computed value as a fraction of the total number of perturbations. This measure ranges in $[0, 1]$; the closest its value is to zero, the greater the reliability of the explanation signal.
\end{itemize}

\subsection{Implementation Details}

Videos are downsampled to $2$ fps and deep feature representations of the frames are obtained by taking the output of the pool5 layer of GoogleNet \cite{7298594}, trained on ImageNet \cite{5206848}. The number of applied perturbations $M$ for producing fragment-level explanations was set equal to $20.000$, in order to have robust and reliable results. The number of applied perturbations $N$ for producing object-level explanations was set equal to $2.000$, as there were only a few visual objects within the selected keyframes and thus the number of possible perturbations was also small. As stated previously, the number of video fragments for producing explanations (both at the fragment and the object level) was set equal to three. For video summarization, we use pre-trained models of the CA-SUM method \cite{10.1145/3512527.3531404} on the SumMe and TVSum datasets. All experiments were carried out on an NVIDIA RTX 4090 GPU card. The utilized models of CA-SUM and the code for reproducing the reported results, will be made publicly available upon acceptance at: \url{https://github.com/IDT-ITI/XAI-Video-Summaries}

\begin{table}[t]
\centering
\caption{Performance of the considered fragment-level explanation methods on the SumMe dataset. The top part shows the computed Disc+/- and SV scores after taking into account videos that have at least one top- and one bottom-scoring fragment identified by the explanation method. The bottom part shows the computed scores after taking into account a smaller subset of the videos, i.e. those that have at least three top- and three bottom-scoring fragments identified by the explanation method. The best scores are shown in bold. The arrows indicate the optimal (lower or higher) value for each evaluation measure.}
\label{tab:fragment-level-summe}
\begin{tabular}{|cccccccc|}
\hline
\multicolumn{2}{|c|}{}                                                        & \multicolumn{1}{c|}{Disc+ ($\downarrow$)} & \multicolumn{1}{c|}{Disc+ Seq ($\downarrow$)} & \multicolumn{1}{c|}{Disc- ($\uparrow$)} & \multicolumn{1}{c|}{Disc- Seq ($\uparrow$)} & \multicolumn{1}{c|}{SV  ($\downarrow$)}    & SV Seq ($\downarrow$)                \\ \hline \hline
\multicolumn{1}{|c|}{\multirow{2}{*}{Top/Bottom-1}} & \multicolumn{1}{c|}{Attention} & \multicolumn{1}{c|}{\textbf{0.568}} & \multicolumn{1}{c|}{-} & \multicolumn{1}{c|}{\textbf{0.971}} & \multicolumn{1}{c|}{-} & \multicolumn{1}{c|}{\textbf{0.063}} & -        \\
\multicolumn{1}{|c|}{}                       & \multicolumn{1}{c|}{LIME}      & \multicolumn{1}{c|}{0.747}          & \multicolumn{1}{c|}{-}          & \multicolumn{1}{c|}{0.886}          & \multicolumn{1}{c|}{-}          & \multicolumn{1}{c|}{0.438}          & -                 \\ \hline \hline
\multicolumn{1}{|c|}{\multirow{2}{*}{Top/Bottom-1}} & \multicolumn{1}{c|}{Attention} & \multicolumn{1}{c|}{\textbf{0.617}} & \multicolumn{1}{c|}{-} & \multicolumn{1}{c|}{\textbf{0.951}} & \multicolumn{1}{c|}{-} & \multicolumn{1}{c|}{\textbf{0.000}}               &         
- \\  \multicolumn{1}{|c|}{}            & \multicolumn{1}{c|}{LIME}      & \multicolumn{1}{c|}{0.879}          & \multicolumn{1}{c|}{-}          & \multicolumn{1}{c|}{0.802}          & \multicolumn{1}{c|}{-}          & \multicolumn{1}{c|}{0.600}               &   -                    \\ \hline
\multicolumn{1}{|c|}{\multirow{2}{*}{Top/Bottom-2}} & \multicolumn{1}{c|}{Attention} & \multicolumn{1}{c|}{\textbf{0.888}} & \multicolumn{1}{c|}{\textbf{0.546}} & \multicolumn{1}{c|}{\textbf{0.980}} & \multicolumn{1}{c|}{\textbf{0.930}} & \multicolumn{1}{c|}{\textbf{0.400}}               &       \textbf{0.200}                \\
\multicolumn{1}{|c|}{}                       & \multicolumn{1}{c|}{LIME}      & \multicolumn{1}{c|}{0.891}          & \multicolumn{1}{c|}{0.785}          & \multicolumn{1}{c|}{0.966}          & \multicolumn{1}{c|}{0.759}          & \multicolumn{1}{c|}{\textbf{0.400}}               &            0.600           \\ \hline
\multicolumn{1}{|c|}{\multirow{2}{*}{Top/Bottom-3}} & \multicolumn{1}{c|}{Attention} & \multicolumn{1}{c|}{0.967}          & \multicolumn{1}{c|}{\textbf{0.547}} & \multicolumn{1}{c|}{\textbf{0.955}} & \multicolumn{1}{c|}{\textbf{0.886}} & \multicolumn{1}{c|}{\textbf{0.400}} & \textbf{0.400}        \\
\multicolumn{1}{|c|}{}                       & \multicolumn{1}{c|}{LIME}      & \multicolumn{1}{c|}{\textbf{0.945}} & \multicolumn{1}{c|}{0.750}          & \multicolumn{1}{c|}{0.918}          & \multicolumn{1}{c|}{0.658}          & \multicolumn{1}{c|}{0.600}          & 0.600                 \\ \hline
\end{tabular}
\end{table}

\subsection{Quantitative Results}

The results about the performance of the examined fragment-level explanation methods on the videos of the SumMe and TVSum datasets, are presented in Tables \ref{tab:fragment-level-summe} and \ref{tab:fragment-level-tvsum}, respectively. In each case, the top part of the Table shows the computed Disc+/- and SV scores after taking into account videos that have at least one top- and one bottom-scoring fragment by the explanation method, while the bottom part shows the computed scores after taking into account videos that have at least three top- and three bottom-scoring fragments by the explanation method. As stated in Section \ref{subsec:data_eval}, the top-scoring fragments are used for computing Disc+ and Disc+ Seq, the bottom-scoring fragments are employed for computing Disc- and Disc- Seq, while both top- and bottom-scoring fragments are utilized for computing SV and SV Seq. For the sake of space, in Tables \ref{tab:fragment-level-summe}-\ref{tab:object-level-tvsum-2} we show the top- and bottom-k scoring fragment (with k equal to 1, 2 or 3) in the same cell. The results in Tables \ref{tab:fragment-level-summe} and \ref{tab:fragment-level-tvsum} show that the attention-based method performs clearly better compared to LIME, in most evaluation settings. The produced fragment-level explanations by this method are more capable to spot the most influential video fragment, while its performance is comparable with that of the LIME-based explanation at spotting the second and third most influential ones; though the attention-based explanations are better at detecting the fragments with the lowest influence (see columns ``Disc+'' and ``Disc-''). The competitiveness of the attention-based method is more pronounced when more than one video fragments are taken into account, as it performs constantly better than LIME in both datasets (see columns ``Disc+ Seq'' and ``Disc- Seq''). Finally, the produced fragment-level explanations by the aforementioned method are clearly more effective in discriminating the most from the least influential fragments of the video, as indicated by the significantly lower sanity violation scores in all settings (see columns ``SV'' and ``SV Seq''). 

\begin{table}[t]
\centering
\caption{Performance of the considered fragment-level explanation methods on the TVSum dataset. The top part shows the computed Disc+/- and SV scores after taking into account videos that have at least one top- and one bottom-scoring fragment identified by the explanation method. The bottom part shows the computed scores after taking into account a smaller subset of the videos, i.e. those that have at least three top- and three bottom-scoring fragments identified by the explanation method. The best scores are shown in bold. The arrows indicate the optimal (lower or higher) value for each evaluation measure.}
\label{tab:fragment-level-tvsum}
\begin{tabular}{|cccccccc|}
\hline
\multicolumn{2}{|c|}{}                                                        & \multicolumn{1}{c|}{Disc+ ($\downarrow$)} & \multicolumn{1}{c|}{Disc+ Seq ($\downarrow$)} & \multicolumn{1}{c|}{Disc- ($\uparrow$)} & \multicolumn{1}{c|}{Disc- Seq ($\uparrow$)} & \multicolumn{1}{c|}{SV  ($\downarrow$)}    & SV Seq ($\downarrow$)                \\ \hline \hline
\multicolumn{1}{|c|}{\multirow{2}{*}{Top/Bottom-1}} & \multicolumn{1}{c|}{Attention} & \multicolumn{1}{c|}{\textbf{0.579}} & \multicolumn{1}{c|}{-} & \multicolumn{1}{c|}{\textbf{0.983}} & \multicolumn{1}{c|}{-} & \multicolumn{1}{c|}{\textbf{0.000}} & -              \\
\multicolumn{1}{|c|}{}                       & \multicolumn{1}{c|}{LIME}      & \multicolumn{1}{c|}{0.798}          & \multicolumn{1}{c|}{-}          & \multicolumn{1}{c|}{0.952}          & \multicolumn{1}{c|}{-}          & \multicolumn{1}{c|}{0.298}          & -                       \\ \hline \hline
\multicolumn{1}{|c|}{\multirow{2}{*}{Top/Bottom-1}} & \multicolumn{1}{c|}{Attention} & \multicolumn{1}{c|}{\textbf{0.561}} & \multicolumn{1}{c|}{-} & \multicolumn{1}{c|}{\textbf{0.984}} & \multicolumn{1}{c|}{-}            & \multicolumn{1}{c|}{\textbf{0.000}}    & -   \\
\multicolumn{1}{|c|}{}                       & \multicolumn{1}{c|}{LIME}      & \multicolumn{1}{c|}{0.795}          & \multicolumn{1}{c|}{-}          & \multicolumn{1}{c|}{0.940}          & \multicolumn{1}{c|}{-}          & \multicolumn{1}{c|}{0.308}               & \multicolumn{1}{c|}{-}       \\ \hline
\multicolumn{1}{|c|}{\multirow{2}{*}{Top/Bottom-2}} & \multicolumn{1}{c|}{Attention} & \multicolumn{1}{c|}{0.967} & \multicolumn{1}{c|}{\textbf{0.519}} & \multicolumn{1}{c|}{\textbf{0.990}} & \multicolumn{1}{c|}{\textbf{0.963}} & \multicolumn{1}{c|}{0.333}               & \multicolumn{1}{c|}{\textbf{0.000}}       \\
\multicolumn{1}{|c|}{}                       & \multicolumn{1}{c|}{LIME}      & \multicolumn{1}{c|}{\textbf{0.909}} & \multicolumn{1}{c|}{0.696}          & \multicolumn{1}{c|}{0.954}          & \multicolumn{1}{c|}{0.875}          & \multicolumn{1}{c|}{\textbf{0.308}}               & \multicolumn{1}{c|}{0.282}       \\ \hline
\multicolumn{1}{|c|}{\multirow{2}{*}{Top/Bottom-3}} & \multicolumn{1}{c|}{Attention} & \multicolumn{1}{c|}{0.964}          & \multicolumn{1}{c|}{\textbf{0.483}} & \multicolumn{1}{c|}{\textbf{0.982}} & \multicolumn{1}{c|}{\textbf{0.943}} & \multicolumn{1}{c|}{\textbf{0.333}} & \textbf{0.026}              \\
\multicolumn{1}{|c|}{}                       & \multicolumn{1}{c|}{LIME}      & \multicolumn{1}{c|}{\textbf{0.960}} & \multicolumn{1}{c|}{0.618}          & \multicolumn{1}{c|}{0.969}          & \multicolumn{1}{c|}{0.834}          & \multicolumn{1}{c|}{0.461}          & 0.333                       \\ \hline
\end{tabular}
\end{table}

\begin{figure}[t]
\centering
\includegraphics[width=\textwidth]{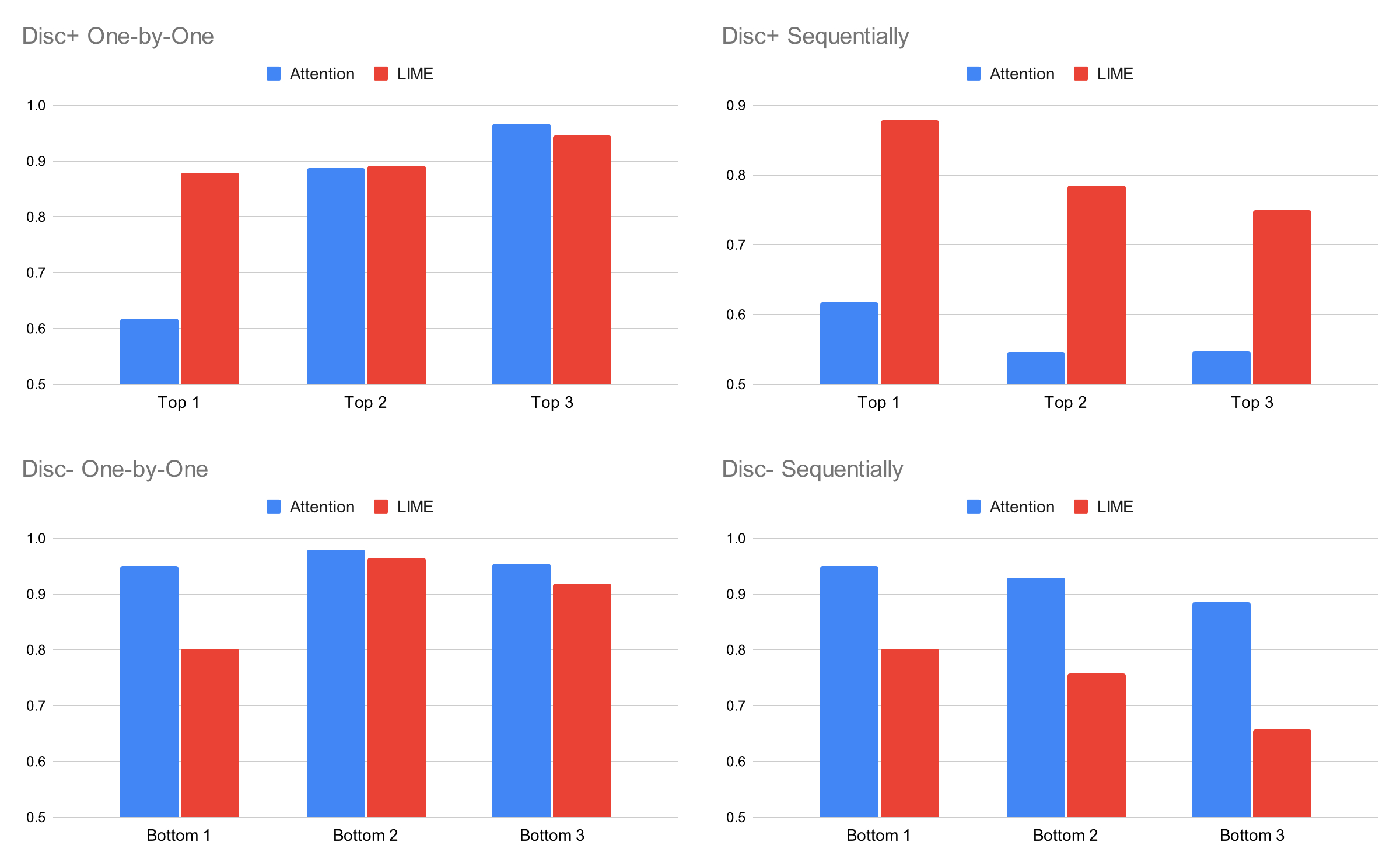}
\caption{The computed Disc+ and Disc- scores for the examined fragment-level explanation methods on the videos on the SumMe dataset, after masking out the three top- and bottom-scoring fragments in a one-by-one and sequential manner.}
\label{fig:top3_summe_fragment}
\end{figure}

\begin{figure}[t]
\centering
\includegraphics[width=\textwidth]{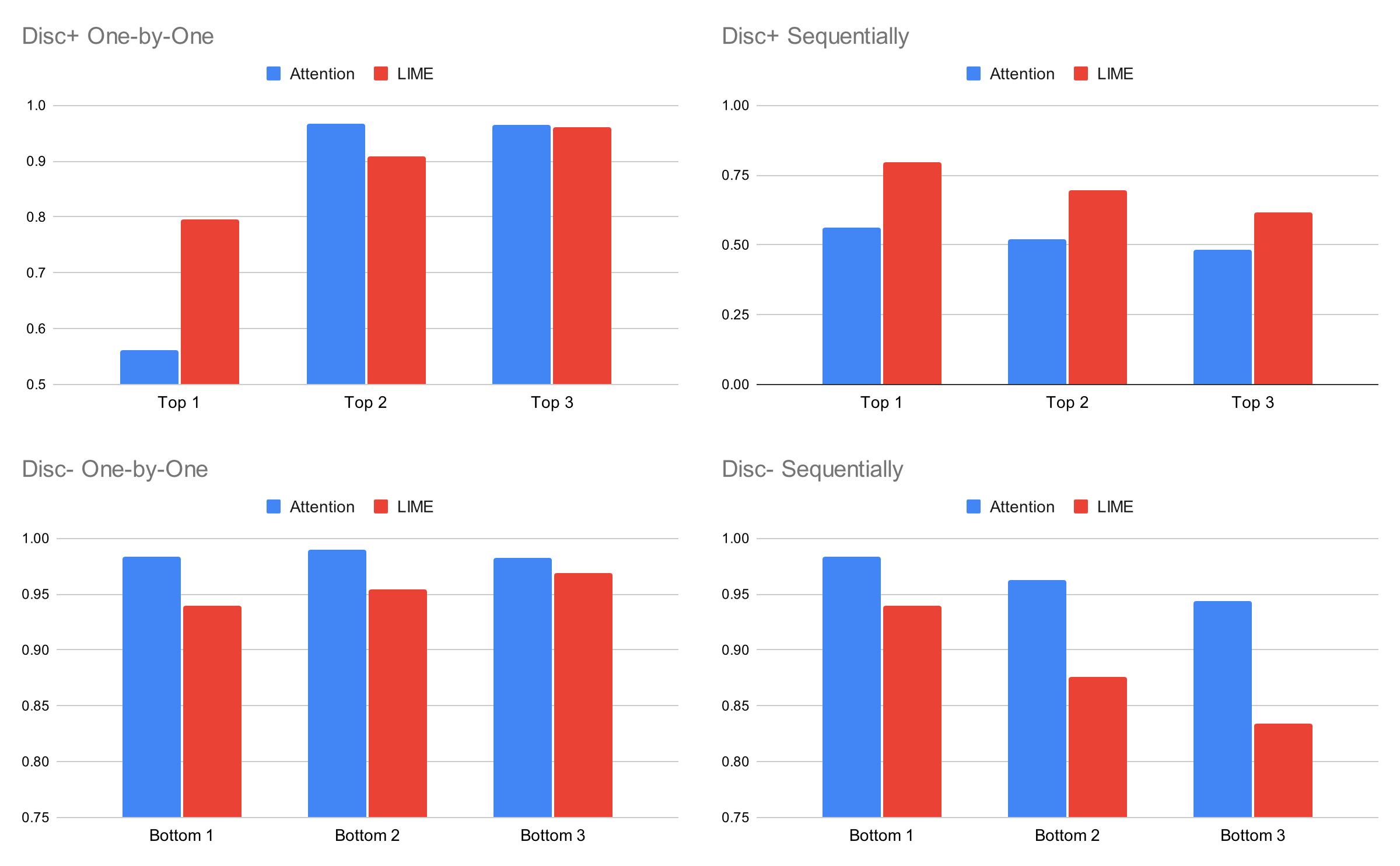}
\caption{The computed Disc+ and Disc- scores for the examined fragment-level explanation methods on the videos on the TVSum dataset, after masking out the three top- and bottom- scoring fragments in a one-by-one and sequential manner.}
\label{fig:top3_tvsum_fragment}
\end{figure}

To assess the competency of the examined fragment-level explanation methods to correctly rank the most and least influential video fragments on the summarization model's output, in Fig. \ref{fig:top3_summe_fragment} and \ref{fig:top3_tvsum_fragment} we illustrate the computed Disc+ and Disc- scores for the videos of the SumMe and TVSum dataset, after masking out the three top- and bottom-scoring fragments, in a one-by-one and sequential manner. The presented scores show, that both methods are able to correctly rank the most influential fragments, as in most cases they lead to Disc+ scores that are gradually increasing when moving from the top-1 to the top-3 scoring fragment (as expected). More specifically, the attention-based method seems to be more appropriate at spotting the fragment with the highest influence to the summarization model (as indicated by the significantly lower Disc+ value for the top-1), while its performance is comparable with the one of LIME when finding the second and third most influential fragment. Moreover, the effectiveness of both methods to rank the most influential fragments is also illustrated by the observed values when masking out these fragments in a sequential manner. The inclusion of additional fragments in the explanation leads to lower Disc+ values (as expected), while the impact of the second and third top-scoring fragments is quantifiable but clearly smaller than the one of the top-1 scoring fragment. Overall, the attention-based explanation method performs better on both datasets, as it leads to significantly lower Disc+ scores compared to LIME (especially on the SumMe dataset). With respect to video fragments that influence the least the output of the summarization model, both methods seem to be less effective at spotting them in the used videos, as the obtained Disc- scores show that, in most cases, the bottom-scoring fragment has a higher impact on the summarization model, compared to the impact of the second and third bottom-scoring fragment (contrary to the expected behavior). Nevertheless, this weakness is less observed for the attention-based method, as the produced explanations lead to similar Disc- scores for the bottom-1, bottom-2 and bottom-3 fragment on both datasets, contrary to the LIME method, which indicates a fragment with clearly higher impact than the other two, as the least influential one (especially on the SumMe dataset). The competitiveness of the attention-based method is also highlighted by the generally higher Disc- scores compared to the ones of the LIME method, after masking out more than one of the least influential video fragments (i.e., sequentially) on the videos of both datasets. ``Disc- Seq'' scores around $0.9$ even after masking out three fragments of the video, point out the competency of this method to spot fragments with minor influence on the output of the summarization model.

\begin{table}[t]
\centering
\caption{Performance of the object-level explanation method on the SumMe dataset using the selected video fragments by the attention-based and LIME explanation methods. The top part shows the computed Disc+/- and SV scores after taking into account videos that have at least one top- and one bottom-scoring visual object identified by the explanation method. The bottom part shows the computed scores after taking into account a smaller subset of the videos, i.e. those that have at least three top- and three bottom-scoring visual objects identified by the explanation method. The best scores are shown in bold. The arrows indicate the optimal (lower or higher) value for each evaluation measure.}
\label{tab:object-level-summe}
\begin{tabular}{|cccccccc|}
\hline
\multicolumn{2}{|c|}{}                                                        & \multicolumn{1}{c|}{Disc+ ($\downarrow$)} & \multicolumn{1}{c|}{Disc+ Seq ($\downarrow$)} & \multicolumn{1}{c|}{Disc- ($\uparrow$)} & \multicolumn{1}{c|}{Disc- Seq ($\uparrow$)} & \multicolumn{1}{c|}{SV  ($\downarrow$)}    & SV Seq ($\downarrow$)                \\ \hline \hline
\multicolumn{1}{|c|}{\multirow{2}{*}{Top/Bottom-1}} & \multicolumn{1}{c|}{Attention} & \multicolumn{1}{c|}{0.969}          & \multicolumn{1}{c|}{-}          & \multicolumn{1}{c|}{\textbf{0.949}} & \multicolumn{1}{c|}{-} & \multicolumn{1}{c|}{0.694}          & -                       \\
\multicolumn{1}{|c|}{}                       & \multicolumn{1}{c|}{LIME}      & \multicolumn{1}{c|}{\textbf{0.941}} & \multicolumn{1}{c|}{-} & \multicolumn{1}{c|}{0.910}          & \multicolumn{1}{c|}{-}          & \multicolumn{1}{c|}{\textbf{0.603}} & -             \\ \hline \hline
\multicolumn{1}{|c|}{\multirow{2}{*}{Top/Bottom-1}} & \multicolumn{1}{c|}{Attention} & \multicolumn{1}{c|}{0.976}          & \multicolumn{1}{c|}{-}          & \multicolumn{1}{c|}{\textbf{0.963}} & \multicolumn{1}{c|}{-} & \multicolumn{1}{c|}{\textbf{0.639}}               & \multicolumn{1}{c|}{-}       \\
\multicolumn{1}{|c|}{}                       & \multicolumn{1}{c|}{LIME}      & \multicolumn{1}{c|}{\textbf{0.937}} & \multicolumn{1}{c|}{-} & \multicolumn{1}{c|}{0.878}          & \multicolumn{1}{c|}{-}          & \multicolumn{1}{c|}{0.666}               & -       \\ \hline
\multicolumn{1}{|c|}{\multirow{2}{*}{Top/Bottom-2}} & \multicolumn{1}{c|}{Attention} & \multicolumn{1}{c|}{0.988}          & \multicolumn{1}{c|}{0.968}          & \multicolumn{1}{c|}{\textbf{0.981}} & \multicolumn{1}{c|}{\textbf{0.958}} & \multicolumn{1}{c|}{\textbf{0.555}}               & \multicolumn{1}{c|}{\textbf{0.639}}       \\
\multicolumn{1}{|c|}{}                       & \multicolumn{1}{c|}{LIME}      & \multicolumn{1}{c|}{\textbf{0.962}} & \multicolumn{1}{c|}{\textbf{0.915}} & \multicolumn{1}{c|}{0.921}          & \multicolumn{1}{c|}{0.839}          & \multicolumn{1}{c|}{0.833}               & \multicolumn{1}{c|}{0.750}       \\ \hline
\multicolumn{1}{|c|}{\multirow{2}{*}{Top/Bottom-3}} & \multicolumn{1}{c|}{Attention} & \multicolumn{1}{c|}{0.994}          & \multicolumn{1}{c|}{0.962}          & \multicolumn{1}{c|}{\textbf{0.989}} & \multicolumn{1}{c|}{\textbf{0.952}} & \multicolumn{1}{c|}{0.750} & \textbf{0.555}              \\
\multicolumn{1}{|c|}{}                       & \multicolumn{1}{c|}{LIME}      & \multicolumn{1}{c|}{\textbf{0.959}} & \multicolumn{1}{c|}{\textbf{0.897}} & \multicolumn{1}{c|}{0.956}          & \multicolumn{1}{c|}{0.828}          & \multicolumn{1}{c|}{\textbf{0.611}}          & 0.805                     \\ \hline
\end{tabular}
\end{table}

\begin{table}[t]
\centering
\caption{Performance of the object-level explanation method on the TVSum dataset using the selected video fragments by the attention-based and LIME explanation methods. The top part shows the computed Disc+/- and SV scores after taking into account videos that have at least one top- and one bottom-scoring visual object identified by the explanation method. The bottom part shows the computed scores after taking into account a smaller subset of the videos, i.e. those that have at least three top- and three bottom-scoring visual objects identified by the explanation method. The best scores are shown in bold. The arrows indicate the optimal (lower or higher) value for each evaluation measure.}
\label{tab:object-level-tvsum}
\begin{tabular}{|cccccccc|}
\hline
\multicolumn{2}{|c|}{}                                                        & \multicolumn{1}{c|}{Disc+ ($\downarrow$)} & \multicolumn{1}{c|}{Disc+ Seq ($\downarrow$)} & \multicolumn{1}{c|}{Disc- ($\uparrow$)} & \multicolumn{1}{c|}{Disc- Seq ($\uparrow$)} & \multicolumn{1}{c|}{SV  ($\downarrow$)}    & SV Seq ($\downarrow$)                \\ \hline \hline
\multicolumn{1}{|c|}{\multirow{2}{*}{Top/Bottom-1}} & \multicolumn{1}{c|}{Attention} & \multicolumn{1}{c|}{0.954}          & \multicolumn{1}{c|}{-}          & \multicolumn{1}{c|}{\textbf{0.989}} & \multicolumn{1}{c|}{-} & \multicolumn{1}{c|}{0.211}          & -        \\
\multicolumn{1}{|c|}{}                       & \multicolumn{1}{c|}{LIME}      & \multicolumn{1}{c|}{\textbf{0.949}} & \multicolumn{1}{c|}{-} & \multicolumn{1}{c|}{0.987}          & \multicolumn{1}{c|}{-}          & \multicolumn{1}{c|}{\textbf{0.162}} & - \\ \hline \hline
\multicolumn{1}{|c|}{\multirow{2}{*}{Top/Bottom-1}} & \multicolumn{1}{c|}{Attention} & \multicolumn{1}{c|}{0.940}          & \multicolumn{1}{c|}{-}          & \multicolumn{1}{c|}{\textbf{0.981}} & \multicolumn{1}{c|}{-} & \multicolumn{1}{c|}{\textbf{0.277}}               &        -        \\
\multicolumn{1}{|c|}{}                       & \multicolumn{1}{c|}{LIME}      & \multicolumn{1}{c|}{\textbf{0.908}} & \multicolumn{1}{c|}{-} & \multicolumn{1}{c|}{0.962}          & \multicolumn{1}{c|}{-}          & \multicolumn{1}{c|}{0.444}               &        -        \\ \hline
\multicolumn{1}{|c|}{\multirow{2}{*}{Top/Bottom-2}} & \multicolumn{1}{c|}{Attention} & \multicolumn{1}{c|}{0.956}          & \multicolumn{1}{c|}{\textbf{0.908}} & \multicolumn{1}{c|}{\textbf{0.995}} & \multicolumn{1}{c|}{\textbf{0.980}} & \multicolumn{1}{c|}{\textbf{0.111}}              &        \textbf{0.111}        \\
\multicolumn{1}{|c|}{}                       & \multicolumn{1}{c|}{LIME}      & \multicolumn{1}{c|}{\textbf{0.948}} & \multicolumn{1}{c|}{0.909}          & \multicolumn{1}{c|}{0.968}          & \multicolumn{1}{c|}{0.907}          & \multicolumn{1}{c|}{0.277}               &      0.611          \\ \hline
\multicolumn{1}{|c|}{\multirow{2}{*}{Top/Bottom-3}} & \multicolumn{1}{c|}{Attention} & \multicolumn{1}{c|}{0.990}          & \multicolumn{1}{c|}{0.889}          & \multicolumn{1}{c|}{\textbf{0.998}} & \multicolumn{1}{c|}{\textbf{0.978}} & \multicolumn{1}{c|}{\textbf{0.111}} & \textbf{0.000} \\
\multicolumn{1}{|c|}{}                       & \multicolumn{1}{c|}{LIME}      & \multicolumn{1}{c|}{\textbf{0.961}} & \multicolumn{1}{c|}{\textbf{0.879}} & \multicolumn{1}{c|}{0.996}          & \multicolumn{1}{c|}{0.907}          & \multicolumn{1}{c|}{\textbf{0.111}}          & 0.500          \\ \hline
\end{tabular}
\end{table}

The performance of the developed method for object-level explanation was initially evaluated using video fragments that were found as the most influential ones by the considered fragment-level explanation methods. The results of our experimental evaluations on the videos of the SumMe and TVSum datasets are presented in Tables \ref{tab:object-level-summe} and \ref{tab:object-level-tvsum}, respectively. These results show that, the object-level explanations for selected video fragments by the two different explanation methods exhibit comparable performance. In general, the LIME-based fragments allow the object-level explanation method to be a bit more effective when spotting the most influential visual objects, while the attention-based fragments lead to better performance when spotting the visual objects with the least influence on the model's output. The comparable capacity of the fragment-level explanation methods is also shown from the mostly similar sanity violation scores. A difference is observed when the applied perturbations affect more than one visual objects, where the produced object-level explanations using the attention-based fragments are associated with clearly lower sanity violation scores. Therefore, a choice between the fragment-level explanation methods could be made based on the level of details in the obtained object-level explanation. If highlighting a single visual object is sufficient, then using the LIME-based fragments could be a good option; however, if the explanation needs to include more visual objects, then the attention-based fragments would be more appropriate for use. In any case, the LIME-based fragment-level explanation method is the only option when there are no details about the video summarization model and thus, the explanation of the model's output must be done through a fully model-agnostic processing pipeline.

The performance of the developed object-level explanation method on the videos of the SumMe and TVSum datasets when using the three top-scoring fragments by the summarization method, is reported in Tables \ref{tab:object-level-summe-2} and \ref{tab:object-level-tvsum-2}, respectively. As a note, in this case, the Disc+/- evaluation measures are computed by taking into account only the importance scores of the frames within the selected fragments. A pair-wise comparison of the Disc+ and Disc- scores shows that our method distinguishes the most from the least influential object in most cases, a fact that is also documented by the obtained sanity violation scores. Moreover, it is able to spot objects that have indeed a very small impact on the output of the summarization process, as demonstrated by the singificantly high Disc- scores. Finally, a cross-dataset comparison shows that our method is more effective on the videos of the TVSum dataset, as it exhibits constantly lower sanity violation scores for both evaluation settings (one-by-one and sequential).

\begin{table}[t]
\centering
\caption{Performance of the object-level explanation method on the SumMe dataset using the selected video fragments by the summarization method. The top part shows the computed Disc+/- and SV scores after taking into account videos that have at least one top- and one bottom-scoring visual object identified by the explanation method. The bottom part shows the computed scores after taking into account a smaller subset of the videos, i.e. those that have at least three top- and three bottom-scoring visual objects identified by the explanation method. The best scores are shown in bold. The arrows indicate the optimal (lower or higher) value for each evaluation measure.}
\label{tab:object-level-summe-2}
\begin{tabular}{|crrrrcc|}
\hline
\multicolumn{1}{|c|}{}                                                        & \multicolumn{1}{c|}{Disc+ ($\downarrow$)} & \multicolumn{1}{c|}{Disc+ Seq ($\downarrow$)} & \multicolumn{1}{c|}{Disc- ($\uparrow$)} & \multicolumn{1}{c|}{Disc- Seq  ($\uparrow$)} & \multicolumn{1}{c|}{SV ($\downarrow$)}    & SV Seq ($\downarrow$)               \\ \hline \hline
\multicolumn{1}{|c|}{Top/Bottom-1} & \multicolumn{1}{c|}{0.894} & \multicolumn{1}{c|}{-}     & \multicolumn{1}{c|}{0.990} & \multicolumn{1}{c|}{-}     & \multicolumn{1}{c|}{0.397} & -  \\ \hline \hline
\multicolumn{1}{|c|}{Top/Bottom-1} & \multicolumn{1}{c|}{0.769} & \multicolumn{1}{c|}{-}     & \multicolumn{1}{c|}{0.977} & \multicolumn{1}{c|}{-}     & \multicolumn{1}{c|}{0.357}      &   -     \\ \hline
\multicolumn{1}{|c|}{Top/Bottom-2} & \multicolumn{1}{c|}{0.985} & \multicolumn{1}{c|}{0.692}     & \multicolumn{1}{c|}{0.995} & \multicolumn{1}{c|}{0.912}     & \multicolumn{1}{c|}{0.365}      &     0.516   \\ \hline
\multicolumn{1}{|c|}{Top/Bottom-3} & \multicolumn{1}{c|}{0.999} & \multicolumn{1}{c|}{0.881}     & \multicolumn{1}{c|}{0.994} & \multicolumn{1}{c|}{0.715}     & \multicolumn{1}{c|}{0.484} & 0.476  \\ \hline
\end{tabular}
\end{table}

\begin{table}[]
\centering
\caption{Performance of the object-level explanation method on the TVSum dataset using the selected video fragments by the summarization method. The top part shows the computed Disc+/- and SV scores after taking into account videos that have at least one top- and one bottom-scoring visual object identified by the explanation method. The bottom part shows the computed scores after taking into account a smaller subset of the videos, i.e. those that have at least three top- and three bottom-scoring visual objects identified by the explanation method. The best scores are shown in bold. The arrows indicate the optimal (lower or higher) value for each evaluation measure.}
\label{tab:object-level-tvsum-2}
\begin{tabular}{|crrrrcc|}
\hline
\multicolumn{1}{|c|}{}                                                        & \multicolumn{1}{c|}{Disc+ ($\downarrow$)} & \multicolumn{1}{c|}{Disc+ Seq ($\downarrow$)} & \multicolumn{1}{c|}{Disc-  ($\uparrow$)} & \multicolumn{1}{c|}{Disc- Seq  ($\uparrow$)} & \multicolumn{1}{c|}{SV ($\downarrow$)}    & SV Seq ($\downarrow$)               \\ \hline \hline
\multicolumn{1}{|c|}{Top/Bottom-1} & \multicolumn{1}{c|}{0.772} & \multicolumn{1}{c|}{-}     & \multicolumn{1}{c|}{0.996} & \multicolumn{1}{c|}{-}     & \multicolumn{1}{c|}{0.195} & -  \\ \hline \hline
\multicolumn{1}{|c|}{Top/Bottom-1} & \multicolumn{1}{c|}{0.883} & \multicolumn{1}{c|}{-}     & \multicolumn{1}{c|}{0.879} & \multicolumn{1}{c|}{-}     & \multicolumn{1}{c|}{0.255}      &   -     \\ \hline
\multicolumn{1}{|c|}{Top/Bottom-2} & \multicolumn{1}{c|}{0.655} & \multicolumn{1}{c|}{0.506}     & \multicolumn{1}{c|}{0.997} & \multicolumn{1}{c|}{0.832}     & \multicolumn{1}{c|}{0.222}      &    0.155   \\ \hline
\multicolumn{1}{|c|}{Top/Bottom-3} & \multicolumn{1}{c|}{0.964} & \multicolumn{1}{c|}{-0.184}    & \multicolumn{1}{c|}{0.999} & \multicolumn{1}{c|}{0.841}     & \multicolumn{1}{c|}{0.344} & 0.133  \\ \hline
\end{tabular}
\end{table}

\subsection{Qualitative Results}

\begin{figure}[t]
\centering
\includegraphics[width=0.94\textwidth]{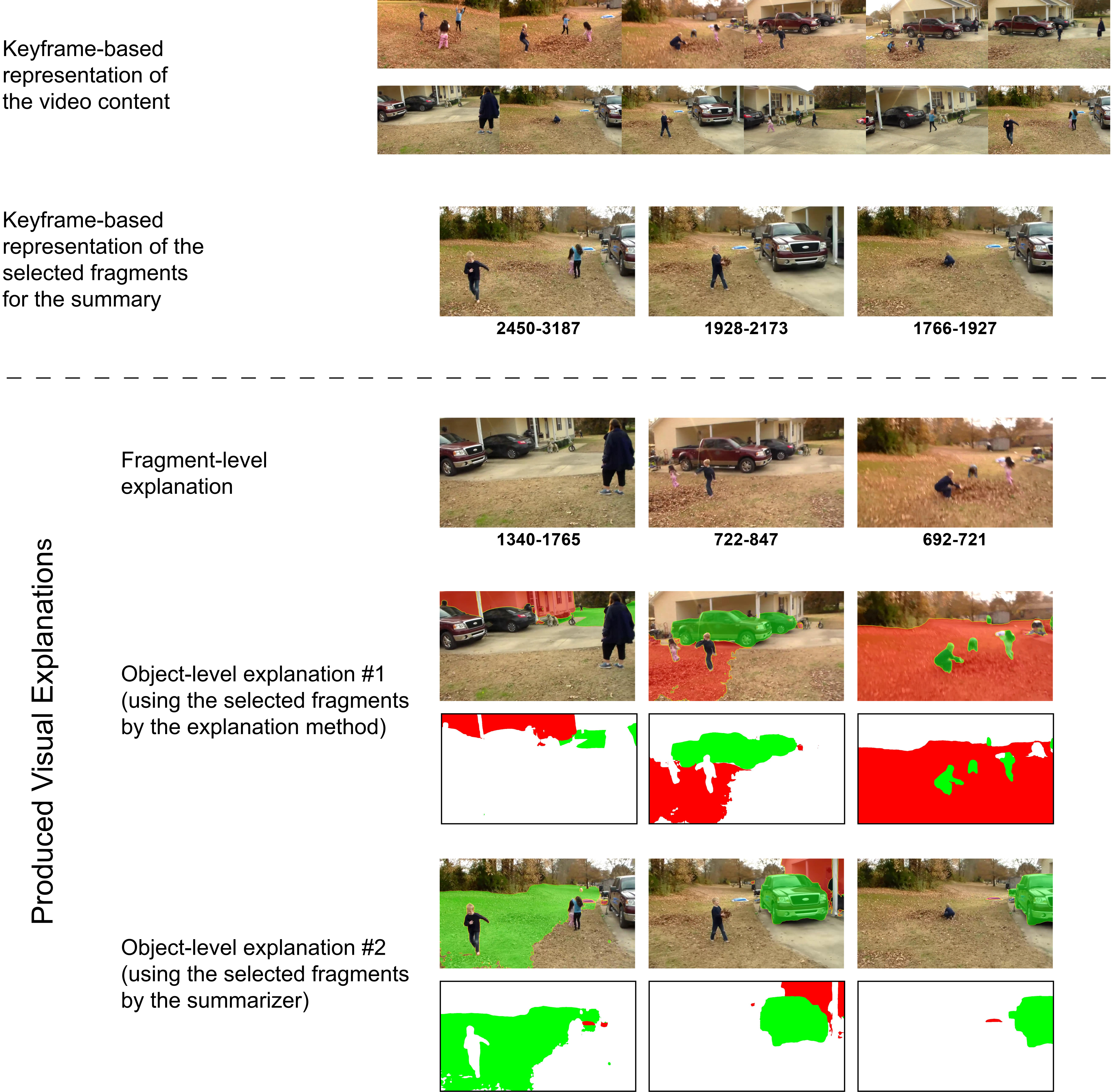}
\caption{Top part: a keyframe-based representation of the visual content of the original and the summarized version of a SumMe video, titled ``Kids playing in leaves''. Bottom part: the produced explanations by the proposed framework. Green- and red-coloured regions in object-level explanations indicate the most and least influential visual objects, respectively.}
\label{fig:summe_qual}
\end{figure}

\begin{figure}[!t]
\centering
\includegraphics[width=\textwidth]{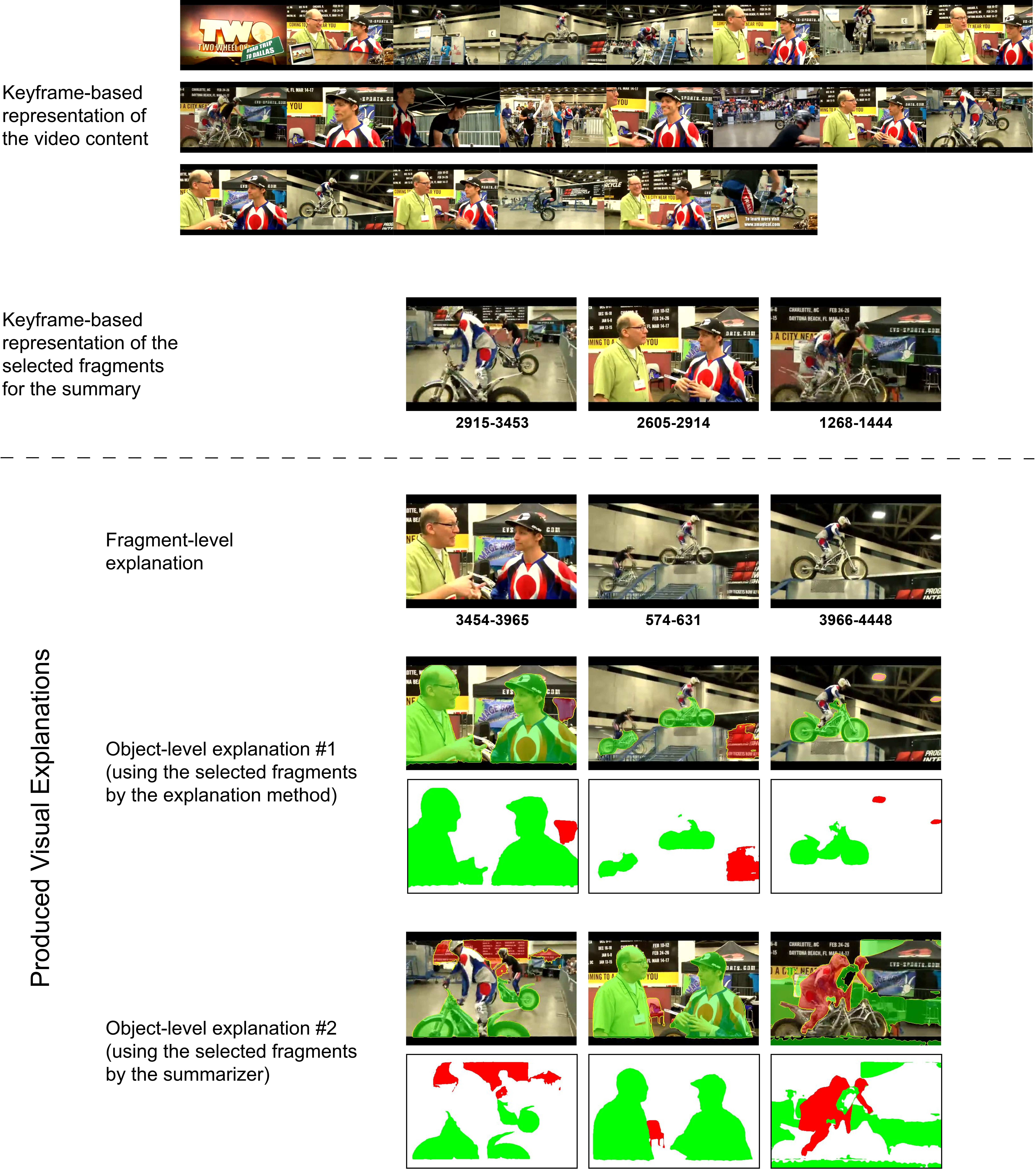}
\caption{Top part: a keyframe-based representation of the visual content of the original and the summarized version of a TVSum video, titled ``Smage Bros. Motorcycle Stunt Show''. Bottom part: the produced explanations by the proposed framework. Green- and red-coloured regions in object-level explanations indicate the most and least influential visual objects, respectively.}
\label{fig:tvsum_qual}
\end{figure}

Our qualitative analysis is based on the produced explanations for two videos of the SumMe and TVSum datasets. The top part of Figs. \ref{fig:summe_qual} and \ref{fig:tvsum_qual} provides a keyframe-based representation of the visual content of the original and the summarized version of the video, while the bottom part shows the produced explanations by the proposed framework. The green- and red-coloured regions in the frames of the object-level explanations, indicate the most and least influential visual objects, respectively. To avoid confusion, these regions are demarcated also in segmentation masks, right below.

In the example video of Fig. \ref{fig:summe_qual}, which is titled ``Kids playing in leaves'', the generated video summary contains parts of the video showing the kids playing with the leaves near a truck. The produced fragment-level explanation from the utilized attention-based method shows that the summarization model paid attention on instances of the kids playing with the leaves (second and third fragment), and the part of the scene where the event is mainly taking place (second fragment). The obtained object-level explanation using the selected fragments by the attention-based explanation method demonstrates that the summarizer concentrates on the truck (second fragment) and the kids (third fragment) - while it pays less attention on the house (first fragment) and the yard (second and third fragment) - thus further explaining why these parts of the video were selected for inclusion in the summary and why other parts of the video (showing the yard right in front of the house, the black car in the parking and the lady) were not. Finally, the produced object-level explanation using the selected fragments by the summarizer seems to partially overlap with the aforementioned one, as it shows that the truck and the house were again the most and least important visual objects for the summarizer (second and third fragment); though, it indicates that the summarizer paid attention to the yard where the kids are playing at.

In the example video of Fig. \ref{fig:tvsum_qual}, which is titled ``Smage Bros. Motorcycle Stunt Show'', the created video summary shows the riders of the motorcycles and one of them being interviewed. The obtained fragment-level explanation from the employed method indicates that the summarizer concentrates on the riders (second and third fragment) and the interview (first fragment). Further insights are given by the object-level explanation of the aforementioned fragments, which demonstrates that the motorcycles (second and third fragment) and the participants in the interview (first fragment) were the most influential visual objects. Similar remarks can be made by observing the produced object-level explanation using the selected fragments from the summarizer (see the first and second fragment). These findings explain why the summarizer selected these parts of the video for inclusion in the summary and why other parts (showing the logo of the TV-show, distant views of the scene and close-ups of the riders) where found as less appropriate. These paradigms show that the produced multi-granular explanations by the proposed framework could allow the user to get insights about the focus of the summarization model, and thus, assist the explanation of the summarization outcome.

\section{Conclusions and Future Steps}

In this paper, we presented a framework for explaining video summarization results through visual-based explanations that are associated with different levels of data granularity. In particular, our framework can provide fragment-level explanations that show the temporal fragments of the video that influenced the most the decisions of the summarizer, using either a model-specific (attention-based) or model-agnostic (LIME-based) explanation method. Moreover, it can produce object-level explanations that highlight the visual objects with the highest influence to the summarizer, taking into account the video fragments that were selected either by the fragment-level explanation method or the summarizer. The performance of the produced explanations was evaluated using a state-of-the-art method (CA-SUM) and two datasets (SumMe and TVSum) for video summarization. The conducted quantitative evaluations showed the effectiveness of our explanation framework to spot the parts of the video (fragments or visual objects) with the highest and lowest influence on the summarizer, while our qualitative analysis demonstrated its capacity to produce a set of multi-granular and informative explanations for the results of the video summarization process. In terms of future steps, we plan to test the performance of our framework using additional state-of-the-art methods for video summarization. Moreover, we aim to leverage advanced vision-language models (e.g., CLIP \cite{Radford2021LearningTV} and BLIP-2 \cite{10.5555/3618408.3619222}) and extend our framework to provide a textual description of the produced visual-based explanations, thus making it more user-friendly for media professionals. 

\section*{Acknowledgments}
This work was supported by the EU Horizon 2020 programme under grant agreement H2020-951911 AI4Media.


\end{document}